\documentclass[times, review, 10pt]{elsarticle}
\usepackage{enumitem}
\usepackage{amsthm}
\usepackage{subcaption}
\usepackage{graphicx}
\newtheorem{theorem}{Theorem}
\usepackage{rotating}
\usepackage{standalone}
\usepackage{multirow}
\usepackage{booktabs}
\usepackage{pgfplotstable}
\usepackage{booktabs}
\usetikzlibrary{positioning, fit, arrows.meta, shapes.geometric, calc}
\usepackage{scalerel,amssymb}

\def\msquare{\mathord{\scalerel*{\Box}{gX}}}

\usepackage{amssymb}
\usepackage{algorithm}
\usepackage{algpseudocode}
\usepackage{amsmath}
\usepackage{hyperref} 
\usepackage{url} 
\usepackage{etoolbox} 

\hypersetup{breaklinks=true} 

\usepackage{siunitx} %
\usepackage{tikz}
\usepackage{tabularray}
\usetikzlibrary{positioning,fit,backgrounds,calc,shapes.geometric}
\usetikzlibrary{positioning,arrows.meta,shapes,fit,backgrounds,calc}
\usepackage{pgfplots}
\pgfplotsset{compat=1.16}
\usepackage{booktabs}  
\usepackage{tabularx}  
\usepackage{array}     
\usetikzlibrary{external}
\pgfplotsset{compat=1.16}

\journal{Pattern Recognition}

\begin{document}
\def\MakeUppercaseUnsupportedInPdfStrings{\scshape}
\begin{frontmatter}



\title{Introducing the Short-Time Fourier Kolmogorov Arnold Network: A Dynamic Graph CNN Approach for Tree Species Classification in 3D Point Clouds}


\author[1]{Said Ohamouddou}
\ead{said\_ohamouddou@um5.ac.ma}
\author[2]{Mohamed Ohamouddou}
\author[1]{Hanaa El Afia}
\author[1]{Abdellatif El Afia}
\author[2]{Rafik Lasri} 
\author[1]{Raddouane Chiheb }

\affiliation[1]{organization={ENSIAS, Mohammed V University},
	city={Rabat},
	postcode={Avenue Mohammed Ben Abdallah Regragui, Madinat Al Irfane, BP 713, Agdal Rabat},
	country={Morocco}}

\affiliation[2]{organization={FPL, Abdelmalek saadi University},
	postcode={Quartier Mhneche II, Avenue 9 Avril B.P.2117}, 
	city={Tetouan},
	country={Morocco}}

\begin{abstract}
	
	Accurate classification of tree species based on Terrestrial Laser Scanning (TLS) and Airborne Laser Scanning (ALS) is essential for biodiversity conservation. While advanced deep learning models for 3D point cloud classification have demonstrated strong performance in this domain, their high complexity often hinders the development of efficient, low-computation architectures. In this paper, we introduce STFT-KAN, a novel Kolmogorov-Arnold network that integrates the Short-Time Fourier Transform (STFT), which can replace the standard linear layer with activation. We implemented STFT-KAN within a lightweight version of DGCNN, called liteDGCNN, to classify tree species using the TLS data. Our experiments show that STFT-KAN outperforms existing KAN variants by effectively balancing model complexity and performance with parameter count reduction, achieving competitive results compared to MLP-based models. Additionally, we evaluated a hybrid architecture that combines MLP in edge convolution with STFT-KAN in other layers, achieving comparable performance to MLP models while reducing the parameter count by 50\% and 75\% compared to other KAN-based variants. Furthermore, we compared our model to leading 3D point cloud learning approaches, demonstrating that STFT-KAN delivers competitive results compared to the state-of-the-art method PointMLP lite with an 87\% reduction in parameter count.
	
\end{abstract}

\begin{keyword}


Kolmogorov-Arnold Network (KAN), STFT, Dynamic Graph CNN, Tree Species Classification, 3D Point Cloud   
\end{keyword}

\end{frontmatter}



\section{Introduction}

\label{sec:introduction}

Forests are vital to planetary ecosystems because of their biodiversity and essential services, such as timber production, climate regulation, carbon sequestration, and habitat preservation \cite{Forests2022}. The accurate classification of tree species is crucial for effective forest management, resource assessment, and biodiversity conservation, leading to significant research attention \cite{arrizza2024terrestrial,michalowska2021review}.

LiDAR data, including Terrestrial Laser Scanning (TLS) and Airborne Laser Scanning (ALS) are widely used to analyze three-dimensional (3D) geometric features, such as leaf morphology, crown structure, and bark surface variations. TLS produces 3D point clouds that represent external surfaces of objects, typically at the ground level, whereas ALS captures 3D data from an aerial perspective, making it suitable for large-scale vegetation analysis. These point clouds provide detailed spatial data for precise measurements and high-resolution visualizations of tree structures \cite{arrizza2024terrestrial,michalowska2021review}.

Classification of tree species from 3D point clouds presents a significant challenge \cite{allen2023tree,arrizza2024terrestrial, michalowska2021review}. Deep learning methodologies have significantly advanced the classification of tree species using LiDAR data, with numerous models achieving notable performance benchmarks. Comparative evaluations highlight the efficacy of these approaches: PointNet++ demonstrated a high classification accuracy (94.4\%) when augmented with RGB data \cite{liuTreeSpeciesClassification2024a}, whereas its application to backpack laser scanning data yielded similarly robust results \cite{liuTreeSpeciesClassification2022f}. LayerNet, which leverages 3D structural feature extraction, attained 92.5\% accuracy \cite{liuTreeSpeciesClassification2021c}, and PointConv outperformed competing architectures in cross-model assessments, with most frameworks exceeding 90\% accuracy thresholds \cite{liuTreeSpeciesClassification2022e}. Recent innovations, such as the dual attention-graph convolution network (DA-GCN) proposed by \cite{liUrbanTreeSpecies2024b}, integrate spectral and structural data to achieve an overall accuracy of 89.80\%, a kappa coefficient of 0.87 and a macro-F1 score of 87.80\%. Complementary approaches include \cite{seidelPredictingTreeSpecies2021b}'s CNN-based method, which employs 2D projections of 3D point clouds to attain 86\% accuracy across seven species, thereby underscoring the versatility of deep learning in addressing tree species classification challenges.

 A common feature of these methods is their reliance on multilayer perceptron (MLP) architectures. The Dynamic Graph Convolutional Neural Network (DGCNN) represents a promising approach based on MLPs; however, these methods often encounter a trade-off between accuracy and computational efficiency and frequently lack interpretability \cite{talaeikhoeiDeepLearningSystematic2023}, limiting their practical applicability in complex 3D forestry studies.

A promising alternative to conventional MLPs is the Kolmogorov-Arnold Network (KAN) \cite{liuKANKolmogorovArnoldNetworks2025b}, based on the Kolmogorov-Arnold theorem \cite{schmidt-hieberKolmogorovArnoldRepresentationTheorem2021,ConstructiveProofKolmogorovs}. The motivation for adopting KAN architectures stems from their integration of learnable activation functions on edges, thereby enhancing the expressive power, computational efficiency, and interpretability \cite{liuKANKolmogorovArnoldNetworks2025b}. By replacing traditional linear weight matrices with spline-based functions, the KAN achieves high accuracy with fewer parameters, leading to faster convergence and superior generalization capabilities in 3D computer vision tasks. As demonstrated by \cite{kashefiPointNetKANPointNet2024b}, the proposed PointNet-KAN replaces MLPs with KAN in PointNet for 3D classification and segmentation tasks, achieving a competitive performance with a simpler architecture. 

To improve the performance of the KAN, researchers have suggested different variations by replacing the original spline function with other types of basis functions. For example, \cite{xuFourierKANGCFFourierKolmogorovArnold2024c} used periodic cosine and sine functions, which are parts of a Fourier series, to create a smoother functional representation. \cite{ssChebyshevPolynomialBasedKolmogorovArnold2024} use Chebyshev polynomials to make training more efficient. 

While direct applications of KAN in forest and tree management remain unexplored in the literature, recent studies have demonstrated their efficacy across diverse environmental domains. In weather forecasting, KAN have enhanced wind nowcasting accuracy at Madeira International Airport \cite{alvesUseKolmogorovArnold2024} and have predicted solar radiation and outdoor temperature with interpretability in Tokyo \cite{gaoRevolutionaryNeuralNetwork2025a}. These studies collectively highlighted KAN' versatility of KAN in addressing environmental challenges through enhanced accuracy, interpretability, and computational efficiency.

Despite these promising results, the KAN-based approaches have limitations in terms of parameter control. While the parameter count can theoretically be increased by augmenting the spline order in spline-based KAN, increasing the polynomial degree in polynomial-based KAN, or stacking additional KAN layers \cite{liuKANKolmogorovArnoldNetworks2025b}, this approach presents challenges when designing minimal architectures that require few parameters. In many cases, even when hyperparameters are selected to minimize the parameter count, the resulting configuration is not optimized compared with MLP-based architectures.

 This challenge makes using KAN in architectures such as DGCNN for 3D point cloud learning of tree species inefficient because it employs at least three distinct MLPs to process and transform input data. The fundamental question persists: How can one use a simple KAN layer as an alternative to MLP and linear layers within such architectures while maintaining a fixed minimal number of parameters?.

To address this challenge and enhance control over the KAN parametrization, we introduce a novel KAN layer based on the Short-Time Fourier Transform (STFT) \cite{SignalEstimationModified}. The STFT-KAN framework enables the utilization of varying window sizes across different frequency ranges, offering a more flexible and efficient representation of underlying signals by capturing nonstationary frequency features.
This improved Fourier-based KAN formulation provides greater control over model parameters, mitigating the overfitting risk while maintaining a performance comparable to that of the original KAN model. STFT-KAN uses linear operations only, applying windowed Fourier transforms with learnable coefficients to extract frequency patterns from input signals, which can provide a more interpretable version of KAN. We implemented STFT-KAN as an alternative to MLP in a lite version  of DGCNN architecture called liteDGCNN for 3D point cloud multiclass classification of tree species. The performance of our approach was evaluated quantitatively by comparing it with MLP-based liteDGCNN, KAN-based liteDGCNN variants, and other state-of-the-art methods of 3D point cloud classification.

Our key contributions are summarized as follows.
\begin{itemize}
	\item We introduce STFT-KAN, a new and efficient KAN layer based on the Short-Time Fourier Transform.
	
	\item We introduce liteDGCNN, a lighter version of DGCNN, which we use for 3D point cloud classification of tree species.
	
	\item We compare the performance of our STFT-KAN based liteDGCNN to several alternatives, including other KAN versions, the standard MLP-based liteDGCNN, and a hybrid model.
	
	\item STFT-KAN based liteDGCNN performs better in terms of accuracy and efficiency compared to most other KAN-based liteDGCNN versions, while achieving competitive accuracy compared to Fourier KAN with a 75\% reduction in parameter count.
	
	\item The proposed hybrid model which combines both MLP and STFT-KAN components, achieves similar results to the MLP-based approach while reducing the parameter count by 50\%.
	
	\item Compared to leading state-of-the-art deep learning methods for 3D point cloud classification, our STFT-KAN-based liteDGCNN achieves accuracy competitive with PointMLP lite, while reducing the parameter count by 89\%.
	
	\item To encourage reproducibility, we have made our code publicly available at  \url{https://github.com/said-ohamouddou/STFT-KAN-liteDGCNN}.
	
\end{itemize}

The remainder of this paper is organized as follows. Section~\ref{sec:background} presents the theoretical background of our topic, such as the spline-based KAN and Fourier-KAN variants and the DGCNN. Section~\ref{sec:stftkan} presents the novel STFT-KAN methodology. Section~\ref{sec:experiment} details the experimental setup used to train the STFT-KAN-based liteDGCNN and other KAN-based and MLP-based liteDGCNN variants for 3D point cloud classification of tree species, with a particular focus on datasets and model configurations. Section~\ref{sec:results} presents the results and conducts a comprehensive performance comparison. Finally, we conclude with a summary of our findings and perspectives for future research in section~\ref{sec:conclusion}.

\section{Background}
\label{sec:background}

\subsection{Dynamic Graph Convolutional Neural Network (DGCNN)}

\label{subsec:dgcnn}

The Dynamic Graph CNN (DGCNN) \cite{wangDynamicGraphCNN2019} extends point-based deep learning methods like PointNet~\cite{qiPointNetDeepLearning2017a} by incorporating local geometric relationships. Instead of treating points independently, DGCNN constructs a $k$-nearest neighbor ($k$-NN) graph in the learned feature space and applies edge convolutions on the connecting edges. 

\paragraph{Edge Convolution (EdgeConv)}
Let $\{\mathbf{x}_1, \ldots, \mathbf{x}_n\} \subseteq \mathbb{R}^F$ be an $F$-dimensional point cloud with size $n$. A directed graph $\mathcal{G}=(\mathcal{V}, \mathcal{E})$ is formed, where each node (point) has edges to its $k$ nearest neighbors in the current feature space. For each edge $(i,j)$, an edge feature 
\begin{equation}
	\mathbf{e}_{ij} \;=\; h_{\boldsymbol{\Theta}}\bigl(\mathbf{x}_i, \mathbf{x}_j\bigr)
\end{equation}
is computed. A common choice is
\begin{equation}
	h_{\boldsymbol{\Theta}}\bigl(\mathbf{x}_i, \mathbf{x}_j\bigr) \;=\; \bar{h}_{\boldsymbol{\Theta}}\bigl(\mathbf{x}_i,\; \mathbf{x}_j - \mathbf{x}_i\bigr),
\end{equation}
where $\bar{h}_{\boldsymbol{\Theta}}$ is typically a MLP capturing both the central point $\mathbf{x}_i$ and the relative vector $\mathbf{x}_j - \mathbf{x}_i$. A symmetric aggregation (e.g., max or sum) is then applied over the edge features for each vertex to obtain the updated point features as follows:
\begin{equation}
	\mathbf{x}'_{i} \;=\; \mathop{\msquare}_{j: (i,j)\in \mathcal{E}} h_{\boldsymbol{\Theta}}(\mathbf{x}_i,\mathbf{x}_j).
\end{equation}

\paragraph{DGCNN for tree Species classification using LiDAR data}
The original DGCNN architecture has been evaluated for 3D point cloud classification in several studies, including \cite{zhouMFFNetMultiscaleFeature2025a, liuTreeSpeciesClassification2022c, diab2022deep}. However, its performance is notably lower when it is applied to more complex datasets, as demonstrated in \cite{puliti2025benchmarking}. Our research focused on a scenario involving deployment on edge devices. Although DGCNN achieves reasonable accuracy, its complexity and large number of parameters make it challenging for edge device applications, where simpler models are preferred. This underscores the need for a more streamlined and efficient version that is suitable for such environments.

\subsection{Kolmogorov-Arnold Network}
\subsubsection{Kolmogorov-Arnold Representation Theorem}\label{subsec:kart}
The KAN framework was inspired by the Kolmogorov-Arnold representation theorem. \cite{schmidt-hieberKolmogorovArnoldRepresentationTheorem2021,ConstructiveProofKolmogorovs} demonstrated that any multivariate continuous function defined on a bounded domain can be expressed as a finite composition of univariate continuous functions combined with the binary operation of addition. More precisely, for a smooth function \( f:[0,1]^n\to\mathbb{R} \), one has
\begin{equation}\label{eq:KART}
	f(x) = f(x_1,\cdots,x_n)=\sum_{q=1}^{2n+1} \Phi_q\left(\sum_{p=1}^n\phi_{q,p}(x_p)\right),
\end{equation}
where \(\phi_{q,p}:[0,1]\to\mathbb{R}\) and \(\Phi_q:\mathbb{R}\to\mathbb{R}\).

KAN authors \cite{liuKANKolmogorovArnoldNetworks2025b} proposed the parameterization of each one-dimensional function using a B-spline curve, wherein the coefficients associated with the local B-spline basis functions are learnable.

Thus, a trainable KAN layer with \(D_{\text{in}}\)-dimensional inputs and \(D_{\text{out}}\)-dimensional outputs is defined as a matrix of one-dimensional functions
\begin{align}
	{\mathbf\Phi}=\{\phi_{q,p}\},\qquad p=1,2,\cdots,n_{\rm in},\qquad q=1,2,\cdots,n_{\rm out},
\end{align}
with each function \(\phi_{q,p}\) parameterized by trainable parameters.

In general, a KAN is constructed with a composition of \(L\) layers. Given an input vector \(x_0\in\mathbb{R}^{n_0}\), the network output is given by
\begin{equation}\label{eq:KAN_forward}
	{\rm KAN}(x) = (\Phi_{L-1}\circ \Phi_{L-2}\circ\cdots\circ \Phi_{1}\circ \Phi_{0})(x).
\end{equation}

\subsubsection{Fourier-KAN}
\label{subsec:fourierkan}

Training spline-based KAN presents greater challenges than MLP \cite{liuKANKolmogorovArnoldNetworks2025b}. A recent study by \cite{xuFourierKANGCFFourierKolmogorovArnold2024c} suggested decomposing complex mathematical functions into simpler nonlinear components by using Fourier-based parameterization \cite{nussbaumerFastFourierTransform1982}. This approach employs a Discrete Fourier Transform (DFT) representation of the form
\begin{equation}
	\Phi_F(\mathbf{x}) = \sum_{i=1}^{D_{\text{in}}} \sum_{k=1}^{G} \left( \cos(k \mathbf{x}_i) \cdot a_{ik} + \sin(k \mathbf{x}_i) \cdot b_{ik} \right),
\end{equation}

It is evident that the univariate continuous functions in the Fourier-KAN framework take the following form:

\begin{equation}
	\phi_f(x) = \sum_{k=0}^{G} \left( a_k \cdot \cos(kx) + b_k \cdot \sin(kx) \right)
\end{equation}

Here, $D_{\text{in}}$ represents the dimensionality of the input feature space and ( $a_{ik}$ ) and ( $b_{ik}$ ) are trainable Fourier coefficients. Hyperparameter $G$, referred to as the grid size, determines the number of frequency terms incorporated in the series expansion for each input dimension. By adjusting $G$, one can strike a balance between the model expressivity and computational efficiency.

Fourier-KAN faces limitations owing to its quadratic parameter scaling with the input dimension \( D_{\text{in}} \), making it inefficient for processing high-dimensional inputs. For an input dimension \( D_{\text{in}} \), output dimension \( D_{\text{out}} \), and grid size \( G \), the per-sample complexity comparison with an MLP is as follows:

\begin{align*}
	\text{Fourier-KAN} &: \mathcal{O}(D_{\text{out}} D_{\text{in}} G) \\
	\text{Linear Layer} &: \mathcal{O}(D_{\text{out}} D_{\text{in}})
\end{align*}

\section{Method: STFT-KAN}
\label{sec:stftkan}
\subsection{Motivation}
Consider a high-dimensional input vector \(\mathbf{x} \in \mathbb{R}^{D_{\text{in}}}\), where \(D_{\text{in}}\) represents input dimensionality. Inspired by previous studies on signal estimation \cite{SignalEstimationModified}, we propose using Short-Time Fourier Transform (STFT) to reduce the parameter complexity to \(\mathcal{O}(D_{\text{out}} \cdot N_w \cdot G)\) by decomposing the input into \(N_w\) localized windows (\(N_w \ll D_{\text{in}}\)).

\begin{theorem}
Consider the function:
\begin{equation}
\phi_{G,w}(\mathbf{x}_w[n]) = w[n] \, \mathbf{x}_w[n] \sum_{k=1}^{G} \left( a_{wk} \cos\left(\frac{2\pi k n}{W}\right) + b_{wk} \sin\left(\frac{2\pi k n}{W}\right) \right),
\end{equation}
where $\mathbf{x}_w[n]$ is the $n$-th element in the $w$-th window of signal $\mathbf{x}$ over $[a,b]$. Assume:
\begin{enumerate}
    \item The windowed signal $f_w(n) = w[n]\mathbf{x}_w[n]$ is bounded: $|f_w(n)| \leq C$ for all $0 \leq n \leq W-1$.
    \item The Fourier coefficients satisfy $\sum_{k=1}^\infty \left(|a_{wk}|^2 + |b_{wk}|^2\right) < \infty$ (square summability).
\end{enumerate}
Subsequently, $\phi_{G,w}$ converges uniformly to a univariate function over $[a,b]$ as $G \to \infty$.
\end{theorem}
\begin{proof}
Let $\Psi_w(\mathbf{x}_w[n])$ denote the limiting function
\begin{equation}
\Psi_w(\mathbf{x}_w[n]) = w[n] \, \mathbf{x}_w[n] \sum_{k=1}^\infty \left( a_{wk} \cos\left(\frac{2\pi k n}{W}\right) + b_{wk} \sin\left(\frac{2\pi k n}{W}\right) \right).
\end{equation}
The approximation error at finite $G$ is
\begin{equation}
|\Psi_w - \phi_{G,w}| = \left| f_w(n) \sum_{k=G+1}^\infty \left( a_{wk} \cos\left(\frac{2\pi k n}{W}\right) + b_{wk} \sin\left(\frac{2\pi k n}{W}\right) \right) \right|.
\end{equation}
By using the boundedness of $f_w(n)$ and $|\cos(\cdot)|, |\sin(\cdot)| \leq 1$,
\begin{equation}
|\Psi_w - \phi_{G,w}| \leq C \sum_{k=G+1}^\infty \left(|a_{wk}| + |b_{wk}|\right).
\end{equation}
To bind the tail sum, we apply the Cauchy-Schwarz inequality to sequences $\{1\}$ and $\{|a_{wk}| + |b_{wk}|\}$:
\begin{equation}
\sum_{k=G+1}^\infty \left(|a_{wk}| + |b_{wk}|\right) \leq \sqrt{\sum_{k=G+1}^\infty 1^2} \sqrt{\sum_{k=G+1}^\infty \left(|a_{wk}| + |b_{wk}|\right)^2}.
\end{equation}
By simplifying $(|a_{wk}| + |b_{wk}|)^2 \leq 2(|a_{wk}|^2 + |b_{wk}|^2)$,
\begin{equation}
\sum_{k=G+1}^\infty \left(|a_{wk}| + |b_{wk}|\right) \leq \sqrt{2 \sum_{k=G+1}^\infty \left(|a_{wk}|^2 + |b_{wk}|^2\right)} \cdot \sqrt{\sum_{k=G+1}^\infty 1^2}.
\end{equation}
According to the square-summability assumption, $\sum_{k=G+1}^\infty \left(|a_{wk}|^2 + |b_{wk}|^2\right) \to 0$ as $G \to \infty$. Hence:
\begin{equation}
\lim_{G \to \infty} |\Psi_w - \phi_{G,w}| \leq \lim_{G \to \infty} C \sqrt{2} \sqrt{\sum_{k=G+1}^\infty \left(|a_{wk}|^2 + |b_{wk}|^2\right)} = 0.
\end{equation}
Thus:
\begin{equation}
\phi_{G,} \text{ converges uniformly to } \Psi_w \text{ over } [a,b] \text{ as } G \to \infty.
\end{equation}
\end{proof}

\subsection{STFT-KAN Layer Formulation}
The KAN basis function of the proposed STFT-KAN layer is defined as
\begin{equation}
\phi_w(\mathbf{x}_w[n]) = w[n] \, \mathbf{x}_w[n] \sum_{k=1}^{G} \left( a_{wk} \cos\left(\frac{2\pi k n}{W}\right) + b_{wk} \sin\left(\frac{2\pi k n}{W}\right) \right)
\end{equation}
\noindent where \(\phi_w\) explicitly depends on \(x_w[n]\) (the input signal at window \(w\) and time index \(n\)), modulated by the window function \(w[n]\) and harmonic coefficients \(a_{wk}, b_{wk}\). To address multiple output dimensions, we define a set of coefficients \(a_{owk}, b_{owk}\) for each output dimension \(o\). The full expression for the \(o\)th output dimension \(\Phi_F^o(\mathbf{x})\) becomes
\begin{equation}
\Phi_F^o(\mathbf{x}) = \sum_{w=1}^{N_w} \sum_{n=0}^{W-1} w[n] \, \mathbf{x}_w[n] \sum_{k=1}^{G} \left( a_{owk} \cos\left(\frac{2\pi k n}{W}\right) + b_{owk} \sin\left(\frac{2\pi k n}{W}\right) \right) + b_{\text{bias}}^o
\end{equation}

\noindent The key parameters in this formulation are:
\begin{itemize}
    \item \(D_{\text{out}}\): Output dimension
    \item \(L\): Total signal length (If necessary, \(\mathbf{x}\) can be padded or truncated to the length \(L\))

    \item \(N_w\): Number of windows (\(= \left\lfloor \frac{L - W}{S} \right\rfloor + 1\)), where \(S\) is the stride between windows.
    
    \item \(G\): Number of frequency bins per window (grid size)
    \item \(W\): Window size (time-domain samples)
    \item \(a_{owk}, b_{owk}\): Trainable coefficients for output dimension \(o\), window \(w\) and frequency \(k\).
    \item  The complete output is a vector \(\Phi_F(\mathbf{x}) = [\Phi_F^1(\mathbf{x}), \Phi_F^2(\mathbf{x}), \ldots, \Phi_F^{D_{\text{out}}}(\mathbf{x})]\)
    \item \(\text{bias}_o\): Optional learnable bias term for output dimension \(o\)
\end{itemize}

\begin{figure}[htbp]
	\centering
	\includegraphics[scale=0.9]{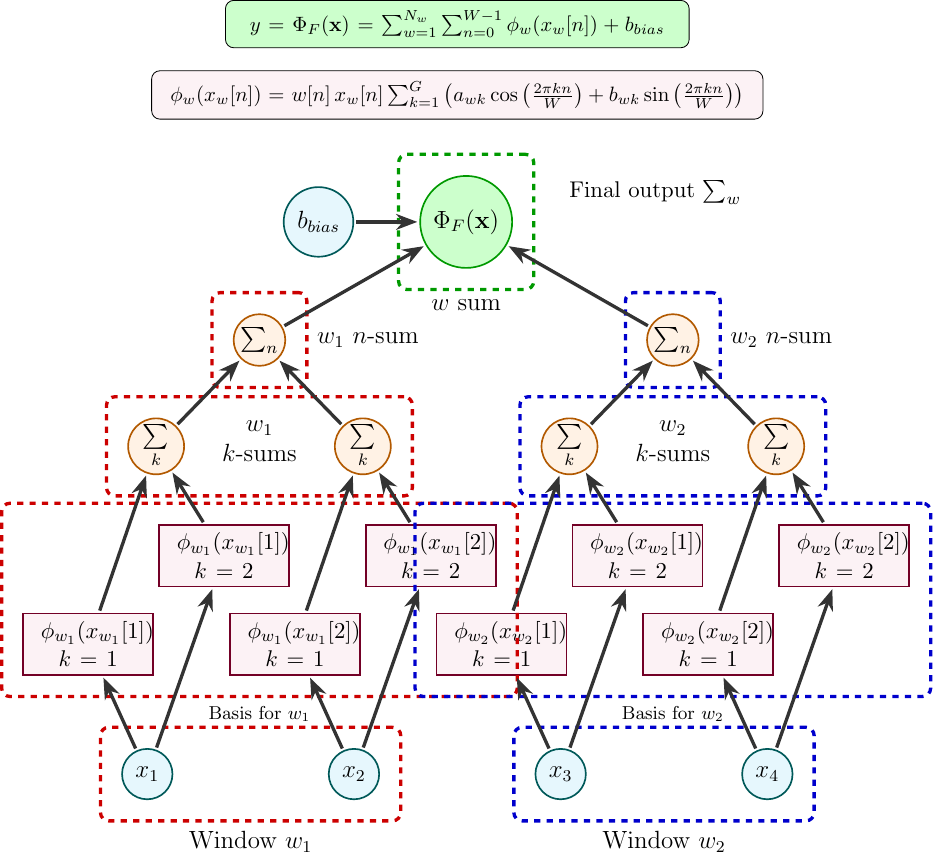}
	\caption{Architecture of a STFT-KAN layer, which takes an input \(\mathbf{x} \in \mathbb{R}^{4}\) and produces a scalar output \(D_{\text{out}} = 1\). The layer operates with a window size \(W = 2\), stride \(S = 2\), and grid size \(G = 2\).}
	\label{fig:stft_arch}
\end{figure}

The STFT-KAN layer projects inputs onto localized Fourier bases, preserving representational capacity while enhancing computational efficiency and interpretability because the Short-Time Fourier Transform (STFT) is a typical linear transform \cite{ParameterisedTimefrequencyAnalysis2019}. Unlike Fourier-KAN, it does not directly apply nonlinear functions to the input but instead relies on projections onto parameterized sinusoidal basis functions. This approach preserves the linearity of the signal representation, ensures mathematical , and avoids distortions introduced by nonlinear activation. In contrast to the simple Fourier transform, STFT can handle nonstationary signals by breaking the signal into shorter segments, where the signal is approximately stationary. Figure~\ref{fig:stft_arch} illustrates the architecture of a simple STFT-KAN layer, whereas Algorithm~\ref{algo:stft-kan} outlines its general implementation. Table~\ref{tab:window_functions} summarizes the common window functions that can be used with STFT-KAN.

\begin{algorithm}
	\caption{The STFT-KAN forward pass for input vector \(\mathbf{x}\), where \(W\) is the window size, \(S\) is the stride, and \(G\) is the grid size. \(D_{\text{out}}\) is the output dimension, and  \(\mathbf{a}\), \(\mathbf{b}\), \(\mathbf{b}_{\text{bias}}\) are the trainable parameters.}
	\label{algo:stft-kan}
		\begin{minipage}{1.1\textwidth} 
			\begin{algorithmic}[1]
				\Function{STFT-KAN}{$\mathbf{x}, \texttt{window\_type}, W, S, G, D_{\text{out}}, \mathbf{a}, \mathbf{b}, \mathbf{b}_{\text{bias}}$}
				\State Calculate number of windows: \(N_w = \left\lfloor \frac{D_{\text{in}} - W}{S} \right\rfloor + 1\)
				\State Generate window function \(\mathbf{h} \in \mathbb{R}^{W}\) based on \(\texttt{window\_type}\)
				
				\State Calculate total signal length for windowing: \(L = (N_w - 1) \times S + W\)
				\State Pad or truncate \(\mathbf{x}\) to length \(L\) if needed
				
				\State Extract windows: \(\mathbf{x}_{\text{windows}} \in \mathbb{R}^{N_w \times W}\) using sliding windows with stride \(S\)
				\State Apply window function: \(\mathbf{x}_{\text{windowed}} = \mathbf{x}_{\text{windows}} \odot \mathbf{h}\)
				
				\State Define features indices: \(\mathbf{n} = [0, 1, \dots, W-1]\)
				\State Define frequency indices: \(\mathbf{k} = [1, 2, \dots, G]\)
				
				\For{window \(w = 0\) to \(N_w-1\)}
				\For{frequency \(k = 1\) to \(G\)}
				\State Compute basis functions:
				\[
				Y_{\cos}[w,k] = \sum_{n=0}^{W-1} \mathbf{x}_{\text{windowed}}[w,n] \cdot \cos\left(\frac{2\pi k n}{W}\right)
				\]
				\[
				Y_{\sin}[w,k] = \sum_{n=0}^{W-1} \mathbf{x}_{\text{windowed}}[w,n] \cdot \sin\left(\frac{2\pi k n}{W}\right)
				\]
				\EndFor
				\EndFor
				
				\For{output dimension \(d = 1\) to \(D_{\text{out}}\)}
				\State Apply learnable weights:
				\[
				\mathbf{y}[d] = \sum_{w=0}^{N_w-1} \sum_{k=1}^{G} \mathbf{a}[d,w,k] \cdot Y_{\cos}[w,k] + \mathbf{b}[d,w,k] \cdot Y_{\sin}[w,k]
				\]
				\EndFor
				
				\State Add bias: \(\mathbf{y} = \mathbf{y} + \mathbf{b}_{\text{bias}}\)
				\State \Return \(\mathbf{y} \in \mathbb{R}^{D_{\text{out}}}\)
				\EndFunction
			\end{algorithmic}
		\end{minipage}
	
\end{algorithm}

\begin{table}[htbp]
	\centering
	\caption{Examples of window functions that we can use with STFT-KAN.}
	\label{tab:window_functions}
	\renewcommand{\arraystretch}{1.5}
	\begin{tabularx}{\textwidth}{>{\raggedright\arraybackslash}p{2.2cm} >{\raggedright\arraybackslash}X >{\raggedright\arraybackslash}X}
		\toprule
		\textbf{Window Type} & \textbf{Function} & \textbf{Description} \\
		\midrule
		Boxcar
		& 
		$h[n] = 1$ & 
	Computationally efficient, and suitable for applications requiring minimal spectral resolution \cite{RectangleelliottHandbookDigitalSignal2013}.  \\
		\addlinespace[0.5em]
		
		Hann 
		& 
		$h[n] = 0.5 \left(1 - \cos\left(\frac{2\pi n}{W-1}\right)\right)$ & 
		Good frequency resolution and moderate sidelobe suppression. Suitable for general-purpose spectral analysis \cite{HanninggharrisUseWindowsHarmonic2005}. \\
		\addlinespace[0.5em]
		
		Hamming 
		& 
		$h[n] = 0.54 - 0.46 \cos\left(\frac{2\pi n}{W-1}\right)$ & 
		Better sidelobe suppression than the Hann window\cite{HammingenochsonProgrammingAnalysisDigital1969}. \\
		\addlinespace[0.5em]
		
		Bartlett 
		& 
		$h[n] = 1 - \left|\frac{2n}{W-1} - 1\right|$ & 
	The triangular shape provides a simple trade-off between the main lobe width and the sidelobe levels \cite{BartletsmithiiiSpectralAudioSignal2011}. \\
		\addlinespace[0.5em]
		
		Blackman 
		& 
		$h[n] = 0.42 - 0.5 \cos\left(\frac{2\pi n}{W-1}\right) + 0.08 \cos\left(\frac{4\pi n}{W-1}\right)$ & 
		Excellent sidelobe suppression makes it suitable for high dynamic-range signals \cite{BlackmanlaiPracticalDigitalSignal2003}.  \\
		\addlinespace[0.5em]
		
		Kaiser 
		& 
		\resizebox{0.3\textwidth}{!}{$h[n] = \frac{I_0\left(\beta \sqrt{1 - \left(\frac{2n}{W-1} - 1\right)^2}\right)}{I_0(\beta)}$}
		where $I_0(x) = \sum_{k=0}^{\infty} \frac{\left(\frac{x}{2}\right)^{2k}}{(k!)^2}$ & 
		Highly customizable via the $\beta$ parameter, allowing precise control over the sidelobe suppression and main-lobee width \cite{sulistyaningsihPerformanceComparisonBlackman2019}.  \\
		\bottomrule
	\end{tabularx}
\end{table}

\paragraph{STFT-KAN smooth initialization}
The STFT-KAN layer supports smooth initialization, meaning that we can initialize coefficients \( (a_{wk}, b_{wk}) \) with a \( \gamma_k = k^{-2} \) decay. This spectral smoothing approach ensures that higher-frequency components contribute less to the output, promoting a smoother function approximation. Starting with spectrally smooth weights can help the model to converge faster during training by providing a better starting point than random initialization.

\section{Experimental Setup}
\label{sec:experiment}
\subsection{Adapted DGCNN Architecture (liteDGCNN)}

\begin{figure}[t]
\centering
\includegraphics[scale=0.5]{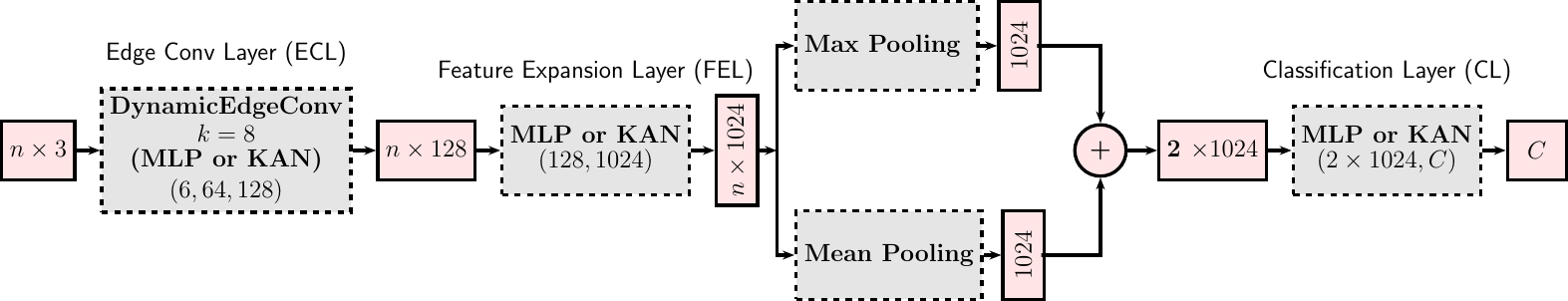}
    \caption{Architecture of liteDGCNN. In the architectural diagram, red blocks represent feature tensors, gray dashed blocks denote processing layers, and arrows illustrate the data flow through the network.}
    \label{fig:dgcnn}
\end{figure}

The adapted architecture of the DGCNN, illustrated in Figure \ref{fig:dgcnn} and referred to as liteDGCNN, processes input data with dimensions $n \times 3$, where $n$ is the number of points. Although more complex architectures could be employed, we opted for a streamlined design in this study to emphasize the efficacy of STFT-KAN and other KAN variants in classifying tree species using an architecture for edge devices. The network operates through the following key stages:

\begin{enumerate} 
	
	\item Edge Convolution Layer (ECL): This layer applies dynamic edge convolution with $k=8$ nearest neighbors. The input features were transformed into a higher-dimensional space using either an MLP or  a KAN with a hidden layer of size $64$, resulting in an output of dimension $n \times 128$.
	
	\item Feature Expansion Layer (FEL): An MLP or KAN is employed to further enhance the feature representation, increasing the dimensionality to $n \times 1024$. The MLP/KAN receives input features of dimension 128 (from edge convolution output) and maps them to $emb\_dims=1024$ dimensions, thereby expanding the feature space.
	
	\item Global Feature Aggregation (GFA): Max pooling and mean pooling are performed in parallel on the expanded features to aggregate local and global geometric information. The outputs are concatenated to capture both the local structural details and broader context of the point cloud.
	
	\item Classification Layer (CL): Finally, a linear layer or a KAN-based layer is used for classification, producing an output of dimension $C$, where $C$ denotes the number of target classes.

\end{enumerate}
Although our architecture uses a single Edge Convolution Layer, and therefore is not dynamic, as the name suggests, we still refer to it as dynamic because it is a special case of the DGCNN.

\subsection{Dataset}
\begin{figure}[h!]
    \centering
    \fbox{\includegraphics[scale=0.2]{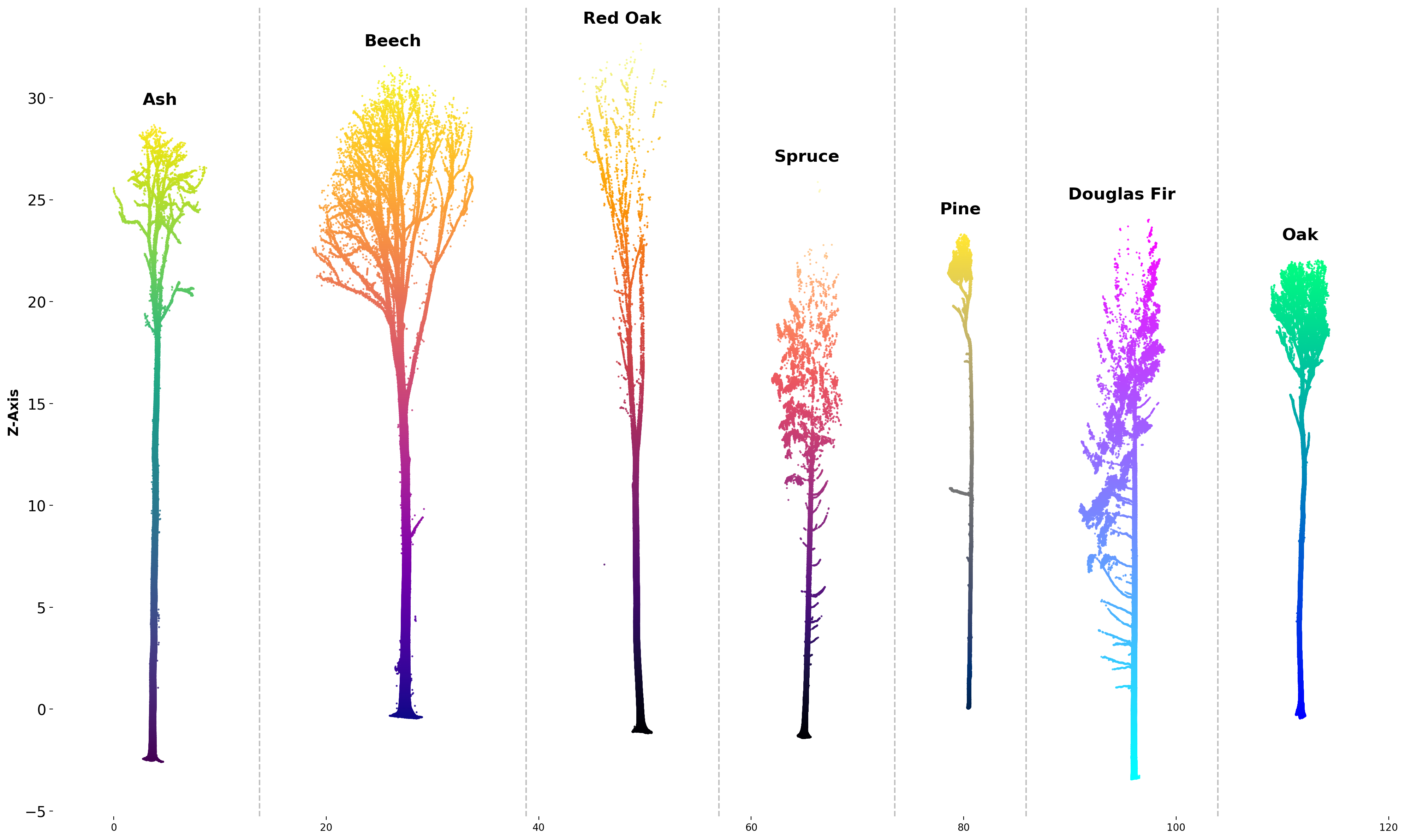}}
    \caption{Sample examples from STPCTLS dataset.}
    \label{fig:sample-examples}
\end{figure}

To objectively evaluate STFT-KAN-based liteDGCNN, we employed publicly available single-tree point clouds from the terrestrial laser scanning (STPCTLS) dataset compiled by \cite{seidelPredictingTreeSpecies2021a}. This dataset encompasses multiple close-range laser scanning measurements of both artificial and natural forests in Germany and the United States, captured using instruments such as the Faro Focus 3-D 120 and Zoller and Fröhlich Imager 5006. It includes seven tree species—beech, red oak, ash, oak, Douglas fir, spruce, and pine—chosen for their pronounced morphological similarity across species and substantial intraspecies variation driven by different growth environments. These factors collectively introduce complexity and necessitate a more robust feature extraction, thereby intensifying the classification challenge. The Table~\ref{tab:stptctls_data} provides the average height, Diameter at Breast Height (DBH), and sample size from the STPTCTLS dataset. Figure~\ref{fig:sample-examples} illustrates the representative examples of individual trees from the STPCTLS dataset.

\begin{table}[h!]
	\centering
	\caption{Tree species summary from the STPTCTLS dataset showing average height, diameter at breast height (DBH), and sample size.}
	\label{tab:stptctls_data}
	\sisetup{table-format=2.2} 
	\begin{tabularx}{\textwidth}{@{}X S[table-format=2.2] S[table-format=1.2] r@{}}
		\toprule
		\textbf{Tree species} & {\textbf{Average height}} & {\textbf{Average DBH}} & \textbf{Sample size} \\
		& {(\si{\meter})} & {(\si{\meter})} &  \\
		\midrule
		Beech        & 27.20 & 0.22 & 164 \\ 
		Douglas fir  & 31.97 & 0.26 & 183 \\ 
		Oak          & 23.82 & 0.22 & 22  \\ 
		Ash          & 30.09 & 0.26 & 39  \\ 
		Spruce       & 26.47 & 0.20 & 158 \\ 
		Pine         & 25.86 & 0.19 & 25  \\ 
		Red oak      & 26.56 & 0.20 & 100 \\ 
		\midrule
		\textbf{Total} & {--} & {--} & \textbf{691} \\ 
		\bottomrule
	\end{tabularx}
\end{table}

\subsection{Data Preprocessing}
We applied a series of preprocessing steps to each 3D point cloud before splitting the dataset into training and testing subsets.

\subsubsection{Point Downsamping}
First, we downsampled the raw scans using a farthest-point sampling technique \cite{eldarFarthestPointStrategy1997} to 1024 points. This method efficiently selects points that preserve the overall shape by maximizing the distance between sampled points, thus retaining the critical geometric details for trees.

 \subsubsection{Normalization}
 Second, we normalized each point cloud to a unit sphere by translating the points to the origin and scaling them based on their furthest distance from the center, which reduces scale discrepancies across various samples and ensures a consistent input for the model. 
\subsubsection{Data Splitting, Balancing and augmentation}

Given the limited sample size, the dataset was partitioned into training and testing subsets using an 80:20 ratio. To maintain consistency, the class distribution was preserved across both the subsets. To mitigate the class imbalance, a class-weighting strategy is incorporated into the loss function, specifically by employing a weighted cross-entropy loss. This strategy identifies minority classes by comparing their sample counts to the mean sample count per class. Classes with sample counts below the mean are assigned proportionally larger weights, ensuring that underrepresented classes are not neglected, while avoiding excessive penalties for extremely small class sizes. Table~\ref{tab:spliting} provides the class distribution for train and test sets. During training, we apply random translations to 3D point clouds to enhance data diversity.

\begin{table}[h!]
  \centering
  \caption{Training and test set distribution.}
  \label{tab:spliting}
  \begin{tabularx}{\textwidth}{l X X X X}
    \toprule
    \multirow{2}{*}{Class} & \multicolumn{2}{c}{\textbf{Training Set}} & \multicolumn{2}{c}{\textbf{Test Set}} \\
    \cmidrule(lr){2-3} \cmidrule(lr){4-5}
     & \textbf{Count} & \textbf{Percentage} & \textbf{Count} & \textbf{Percentage} \\
    \midrule
    Douglas Fir & 146 & 26.45\% & 37 & 26.62\% \\
    Beech       & 131 & 23.73\% & 33 & 23.74\% \\
    Spruce      & 126 & 22.83\% & 32 & 23.02\% \\
    Red Oak     & 80  & 14.49\% & 20 & 14.39\% \\
    Ash         & 31  & 5.62\%  & 8  & 5.76\%  \\
    Pine        & 20  & 3.62\%  & 5  & 3.60\%  \\
    Oak         & 18  & 3.26\%  & 4  & 2.88\%  \\
    \midrule
    \textbf{Total samples} & \multicolumn{2}{r}{\textbf{552}} & \multicolumn{2}{r}{\textbf{139}} \\
    \bottomrule
  \end{tabularx}
\end{table}

\subsection{Experiment Methodology and Implementation Details}

\subsubsection{Layer-by-layer fair comparisons}

In this experiment, we conducted fair layer-by-layer comparisons using the same liteDGCNN architecture. We compared the STFT-KAN-based liteDGCNN with various models, including the traditional MLP-based liteDGCNN and other KAN-based versions such as Spline-KAN \cite{liuKANKolmogorovArnoldNetworks2025b}, Fourier-KAN \cite{xuFourierKANGCFFourierKolmogorovArnold2024c} , FastKAN(RBF-KAN) \cite{liKolmogorovArnoldNetworksAre2024a}, Chebyshev-KAN \cite{ssChebyshevPolynomialBasedKolmogorovArnold2024}, GRAM-Chebyshev-KAN \cite{gramkangithub}, KAL-Net (Legendre Polynomial-KAN) \cite{chen2025legendrekan}, ReLU-KAN \cite{qiuReLUKANNewKolmogorovArnold2024} based liteDGCNN. For the STFT-KAN-based liteDGCNN, a simple grid search was employed to determine the hyperparameters, thereby avoiding extensive tuning. The hyperparameters are summarized in Table~\ref{tab:stftkanminimal}. For the other KAN variants, hyperparameters were selected to minimize the parameter count to the smallest possible value, as detailed in Table~\ref{tab:kanmethodsminimal}. ReLU activation was used for the MLP-based liteDGCNN model. We also tested a combined architecture that used MLP in the edge convolution layer and STFT-KAN in other layers with the same hyperparameters as those previously used in layers other than the grid size in the final layer, which we set to $7$.

\begin{table}[htbp]
	\centering
	\caption{Minimal configuration of STFT-KAN layers: $G$ represents the grid size, $W$ is the window size, $S$ is the stride, SI stands for Smooth Initialization, and WT denotes the window type. The hyperparameters in parentheses are those used in the modified parameters within the combined architecture.}
	\label{tab:stftkanminimal}
	\begin{tabularx}{\textwidth}{@{}lXXXXXX@{}}
		\toprule
		\textbf{Layer} & $G$ & $W$ & $S$ & SI & WT \\ 
		\midrule
		EdgeConv Layer: STFT-KAN 1: & 3 & 2  & 2  & True  & Boxcar \\
		EdgeConv Layer: STFT-KAN 2: & 1 & 28 & 5  & False & Blackman \\
		Embedding Layer:           & 7 & 52 & 20 & True  & Bartlett \\
		Classification Layer:      & 6 & 197& 10 (7) & False & Hann \\
		\bottomrule
	\end{tabularx}
\end{table}

\begin{table}[htbp]
	\centering
	\caption{Minimal configuration for other used KAN variants.}
	\label{tab:kanmethodsminimal}
	\begin{tabular}{@{}lp{4cm}@{}}
		\toprule
		\textbf{Method} & \textbf{Configuration} \\
		\midrule
		Spline-KAN          & Grid Size = 1, Spline Order = 0 \\
		Fourier-KAN         & Grid Size = 1 \\
		FastKAN (RBF-KAN)   & Grid Min = 0, Grid Max = 1, Number of Grids = 2 \\
		Chebyshev-KAN       & Degree = 0 \\
		KAL-Net (Legendre KAN) & Degree = 0 \\
		GRAM-Chebyshev-KAN  & Degree = 0 \\
		ReLU-KAN            & Number of Grids \( g = 1 \), Span Parameter \( k = 0 \) \\
		\bottomrule
	\end{tabular}
\end{table}

\subsubsection{Comparison with State-of-the-Art Models}
In this experiment, we compared the proposed STFT-KAN-based liteDGCNN and hybrid liteDGCNN architecture, which uses both STFT-KAN and MLP, with 14 state-of-the-art models that have demonstrated considerable performance on the ModelNet40 dataset. These models include the full DGCNN and PointNet, PointNet++ SSG \cite{qiPointnetDeepHierarchical2017}, PointNet++ MSG \cite{qiPointnetDeepHierarchical2017}, PointConv \cite{wuPointconvDeepConvolutional2019}, PointMLP, PointMLP Lite \cite{maRethinkingNetworkDesign2022}, PointWEB \cite{zhaoPointWebEnhancingLocal2019}, GDANet \cite{xuLearningGeometryDisentangledRepresentation2021}, PCT \cite{guoPCTPointCloud2021}, PVT \cite{zhangPVTPointVoxelTransformer2022}, CurveNet \cite{muzahidCurveNetCurvaturebasedMultitask2020}, DeepGCN \cite{liDeepGCNsCanGCNs2019}, and PoinTnT \cite{bergPointsPatchesEnabling2022}. For all architectures, we used the default hyperparameters proposed by the authors, except for graph-based architectures, where we set the number of neighbors to $k = 8$, similar to our proposed models.

\subsubsection{Hyperparameter Analysis in STFT-KAN}

 We performed Bayesian Optimization using a Tree-structured Parzen Estimator (TPE) for all STFT-KAN hyperparameters in liteDGCNN over $20$ trials with random initial parameters. During each trial, the model was trained for 300 epochs. This method helps us understand how hyperparameters affect the performance of STFT-KAN and their relationships with each other. Table \ref{tab:bo_bounds} lists the search space bounds for hyperparameters. We used a Functional ANOVA Framework to determine the importance of each hyperparameter.

\begin{table}[ht]
	\centering
	\caption{Bounds of the hyperparameter search space.}
	\begin{tabular}{@{}l l l l@{}}
		\toprule
		\multicolumn{2}{c}{\textbf{ECL: STFT-KAN layer 1}} & \multicolumn{2}{c}{\textbf{ECL: STFT-KAN layer 2}} \\ \midrule
		Grid Size & 1 to 4 & Grid Size & 1 to 7 \\
		Window Size & 2 to 4 & Window Size & 10 to 64 \\
		Stride & 1 to 3 & Stride & 5 to 20 \\
		Smooth Init & True, False & Smooth Init & True, False \\
		Window Type & (boxcar, hann, ...) & Window Type & (boxcar, hann, ...) \\ \midrule
		
		\multicolumn{2}{c}{\textbf{FEL}} & \multicolumn{2}{c}{\textbf{CL}} \\ \midrule
		Grid Size & 5 to 10 & Grid Size & 5 to 8 \\
		Window Size & 20 to 100 & Window Size & 150 to 400 \\
		Stride & 10 to 25 & Stride & 8 to 15 \\
		Smooth Init & True, False & Smooth Init & True, False \\
		Window Type & (boxcar, hann, ...) & Window Type & (boxcar, hann, ...) \\ \bottomrule
	\end{tabular}
	
	\label{tab:bo_bounds}
\end{table}

\subsection{Evaluation Metrics}
To assess the performance of our multiclass classification models, the following metrics were considered:
\begin{itemize}[noitemsep]
	\item Overall Accuracy (OA): Measures the proportion of correct predictions over the total number of predictions:
	\[
	OA = \frac{\text{Number of Correct Predictions}}{\text{Total Number of Predictions}}
	\]
	
	\item Balanced Accuracy (BA): Accounts for class imbalance by averaging the recall for each class:
	\[
	BA = \frac{1}{C} \sum_{c=1}^{C} \frac{\text{TP}_c}{\text{TP}_c + \text{FN}_c}
	\]
	where \( C \) denotes the number of classes, \( \text{TP}_c \) denotes the true positive for class \( c \), and \( \text{FN}_c \) denotes the false negative for class \( c \).
	\item We also use epoch execution time (ET) in seconds (s) and parameter count in millions (M).
\end{itemize}

\subsection{Common Hyperparameters and Training Environment}
All experiments were conducted using the following tools and hardware.
\begin{itemize}
	\item Software: Open3D for data processing, PyTorch and PyTorch Geometric (CUDA 11.8) for modeling, and Scikit-learn for metrics.
	\item Hardware: NVIDIA GeForce RTX 4060 Ti (16 GB VRAM), 32 GB RAM, and an AMD Ryzen 7 5700X 8-core processor. Ubuntu 22.04 LTS.
\end{itemize}

The training hyperparameters listed in Table~\ref{tab:commun-hyperparameters}.

\begin{table}[htbp]
	\centering
	\caption{Common hyperparameter settings for all experiments.}
	\label{tab:commun-hyperparameters}
	\begin{tabular}{lc}
		\toprule
		\textbf{Hyperparameter} & \textbf{Value} \\ 
		\midrule
		Batch Size              & 16 (2 for Fourier-KAN)          \\
		Learning Rate           & 0.001         \\
		Weight Decay            & 0.0001        \\
		Optimizer               & Adam          \\
		Epochs                  & 300           \\
		Learning Rate Scheduler & CosineAnnealingLR ($\eta_{\text{min}} = 1e^{-3}$) \\
		Embedding Dimensions    & 1024          \\
		Aggregation Method      & Max           \\
		Number of Points        & 1024          \\
		Neighbors Number \( k \) & 8           \\
		\bottomrule
	\end{tabular}
\end{table}

\section{Results and Discussion}
\label{sec:results}

\subsection{Layer-by-Layer Comparison}

In this section, we compare the STFT-KAN-based liteDGCNN model and the combined model (which integrates both MLP and STFT-KAN within liteDGCNN) with the MLP-based liteDGCNN and other KAN-based liteDGCNN architectures for 3D point cloud classification of tree species using the STPCTLS dataset. For simplicity, we refer to a liteDGCNN architecture by the name of the layer used, and STFT-KAN-MLP refers to the combined model. Table~\ref{tab:model-comparison-1} presents a summary of the performances of these models by evaluating their overall accuracy, balanced accuracy, parameter count, and epoch execution time.

STFT-KAN performed well in classifying tree species from 3D point clouds, achieving an overall accuracy of 77.70\% and a balanced accuracy of 67.61\%, with a low parameter count of 0.08M. It outperformed all other KAN models, except Fourier-KAN, which achieved an overall accuracy of 79.14\%.

Compared to the majority of the tested KAN models, STFT-KAN reduced the number of parameters by 75\% while maintaining classification performance. However, STFT-KAN had a slightly longer execution time (6.31 s) than the other models, except for ReLU KAN and Fourier-KAN, which required more execution time.

STFT-KAN-MLP, which combines MLP in the edge convolution layer and STFT-KAN in the feature expansion and classification layers, outperforms STFT-KAN in terms of accuracy, achieving an overall accuracy of 85.61\% and a balanced accuracy of 79.02\%, using 0.09M parameters. The utilization of MLP in the edge convolution layer improves the performance, and the execution time decreases to 4.42 seconds. Although traditional MLP models have the highest accuracy (86.33\%) and fastest inference time (1.12 s), they tend to require significantly more parameters (0.16M), whereas STFT-KAN-MLP achieves competitive accuracy with fewer parameters, reducing the parameter count by 50\%.

The incorporation of MLP in edge convolution in the STFT-KAN-based liteDGCNN allows it to outperform all other KAN-based architectures, which often suffer from lower accuracy due to their limited feature extraction capabilities.

\begin{table}[htbp]
	\small
	\centering
	\caption{liteDGCNN performance when using MLP, STFT-KAN, both STFT-KAN and MLP, and other KAN variants.}
	\label{tab:model-comparison-1}
	\begin{tblr}{
			width = \linewidth,
			colspec = {Q[l,m] Q[r,m] Q[r,m] Q[r,m] Q[r,m]},
			row{1} = {font=\bfseries},
			hline{1,2,Z} = {1pt, solid}
		}
		Model & {OA (\%)} & {BA (\%)} & {ET (s)} & {PC (M)} \\
		MLP-LiteDGCNN & \textbf{86.33} & \textbf{80.51} & \textbf{1.12} & 0.16 \\
		\textbf{KAN-LiteDGCNN:} & ~ & ~ & ~ & ~ \\
		~~~FastKAN & 71.94 & 61.29 & 1.27 & 0.31 \\
		~~~Fourier-KAN & 79.14 & 73.20 & 23.94 & 0.31 \\
		~~~Chebyshev-KAN & 59.71 & 51.12 & 1.31 & 0.31 \\
		~~~ReLU-KAN & 71.94 & 59.56 & 7.12 & 0.16 \\
		~~~Gram-Chebyshev-KAN & 69.06 & 58.35 & 1.38 & 0.31 \\
		~~~KAL-Net & 57.55 & 43.80 & 1.38 & 0.31 \\
		~~~Spline-KAN & 71.94 & 62.55 & 1.37 & 0.31 \\
		\textbf{~~~STFT-KAN (our)} & \textbf{77.70} & 67.61 & 6.31 & \textbf{0.08} \\
		\textbf{~~~STFT-KAN-MLP (our)} & \textbf{85.61} & 79.02 & 4.42 & \textbf{0.09} \\
	\end{tblr}
\end{table}

\subsection{Comparison with Existing Architectures}

In this section, we compare LiteDGCNN using the proposed STFT-KAN and STFT-KAN-MLP models with the existing architectures from the literature. Table \ref{tab:model-comparison-2} summarizes the performance of these models.

Point-MLP achieved the highest overall accuracy in our study (89.93\%) but with a large number of parameters (19.56M). Compared to STFT-KAN-MLP, PointMLP Lite achieved a slightly higher overall accuracy of 88.49\% and a balanced accuracy of 81.68\%. However, the STFT-KAN-MLP stands out for using significantly fewer parameters—0.09M compared to 0.71M in PointMLP Lite (an 87\% reduction). This highlights the effectiveness of STFT-KAN-MLP in delivering a competitive performance with a much lighter model architecture.

Similarly, the STFT-KAN model achieved a notable overall accuracy, although it was lower than those of models such as DGCNN (88.49\%) and DeepGCN (87.05\%). It outperformed models such as PointNet++ SSG, PointWEB, PCT, and PointTnT in terms of both accuracy and efficiency. Additionally, STFT-KAN-MLP surpasses more models such as PointNet++ SSG, PVT, and Curvent.

\begin{table}[htbp]
	\small
	\centering
	\caption{State-of-the-art model performance.}
	\label{tab:model-comparison-2}
	\begin{tblr}{
			width = \linewidth,
			colspec = {Q[l,m] Q[l,m] Q[r,m] Q[r,m] Q[r,m] Q[r,m]},
			row{1} = {font=\bfseries},
			hline{1,2,Z} = {1pt, solid},
			hline{9,12} = {0.75pt, solid},
		}
		Category & Model & {OA (\%)} & {BA (\%)} & {ET (s)} & {PC (M)} \\
		MLP & PointNet & 76.98 & 71.51 & \textbf{0.63} & \textbf{0.68} \\
		~ & PointNet++ SSG & 73.38 & 58.95 & 3.07 & 1.46 \\
		~ & PointNet++ MSG & 79.14 & 67.23 & 9.26 & 1.73 \\
		~ & PointConv & 87.77 & 83.14 & 7.97 & 19.56 \\
		~ & PointMLP Lite & \textbf{88.49} & \textbf{81.68} & 3.69 & \textbf{0.71} \\
		~ & PointMLP & \textbf{89.93} & \textbf{89.81} & 13.92 & 13.23 \\
		~ & PointWEB & 73.38 & 71.12 & 24.99 & 0.78 \\
		Attention & GDANet & 87.05 & 83.43 & 7.65 & 0.93 \\
		~ & PCT & 66.19 & 60.14 & 3.84 & 2.87 \\
		~ & PVT & 84.17 & 77.24 & 17.45 & 9.16 \\
		Graph & CurveNet & 84.89 & 78.26 & 5.34 & 2.12 \\
		~ & DeepGCN & 87.05 & 83.72 & 7.89 & 2.21 \\
		~ & PoinTnT & 61.15 & 51.22 & 3.74 & 3.93 \\
		~ & DGCNN & 88.49 & 80.89 & 2.89 & 1.80 \\
	\end{tblr}
\end{table}

\subsection{Parameter Efficiency Analysis}

\begin{figure}[htbp]
	\centering
	\begin{subfigure}[b]{0.4\textwidth}
		\includegraphics[width=\textwidth]{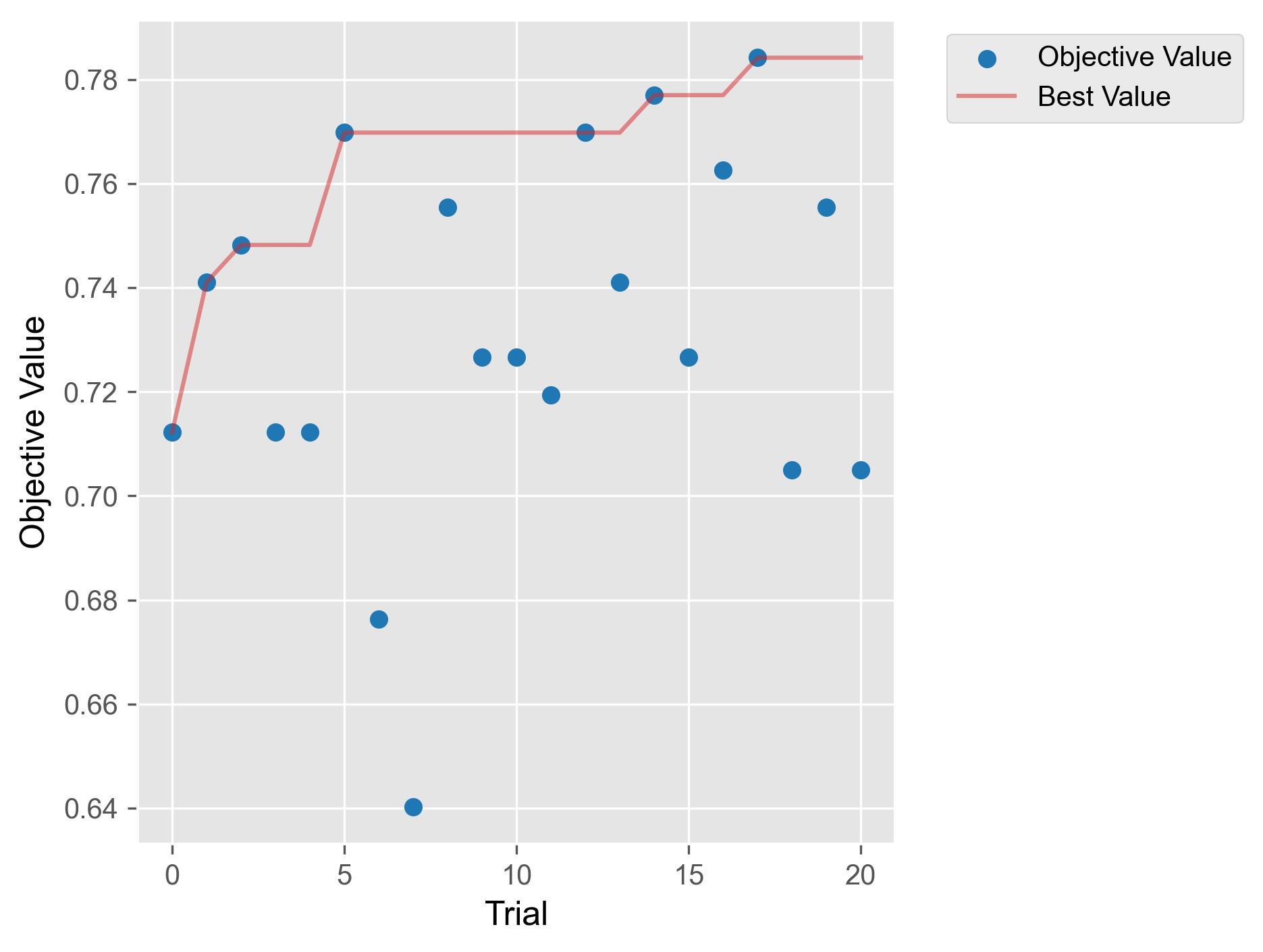}
		\caption{Bayesian optimization history of the STFT-KAN-based liteDGCNN}
		\label{fig:opt_history}
	\end{subfigure}
	\hfill
	\begin{subfigure}[b]{0.4\textwidth}
		\includegraphics[width=\textwidth]{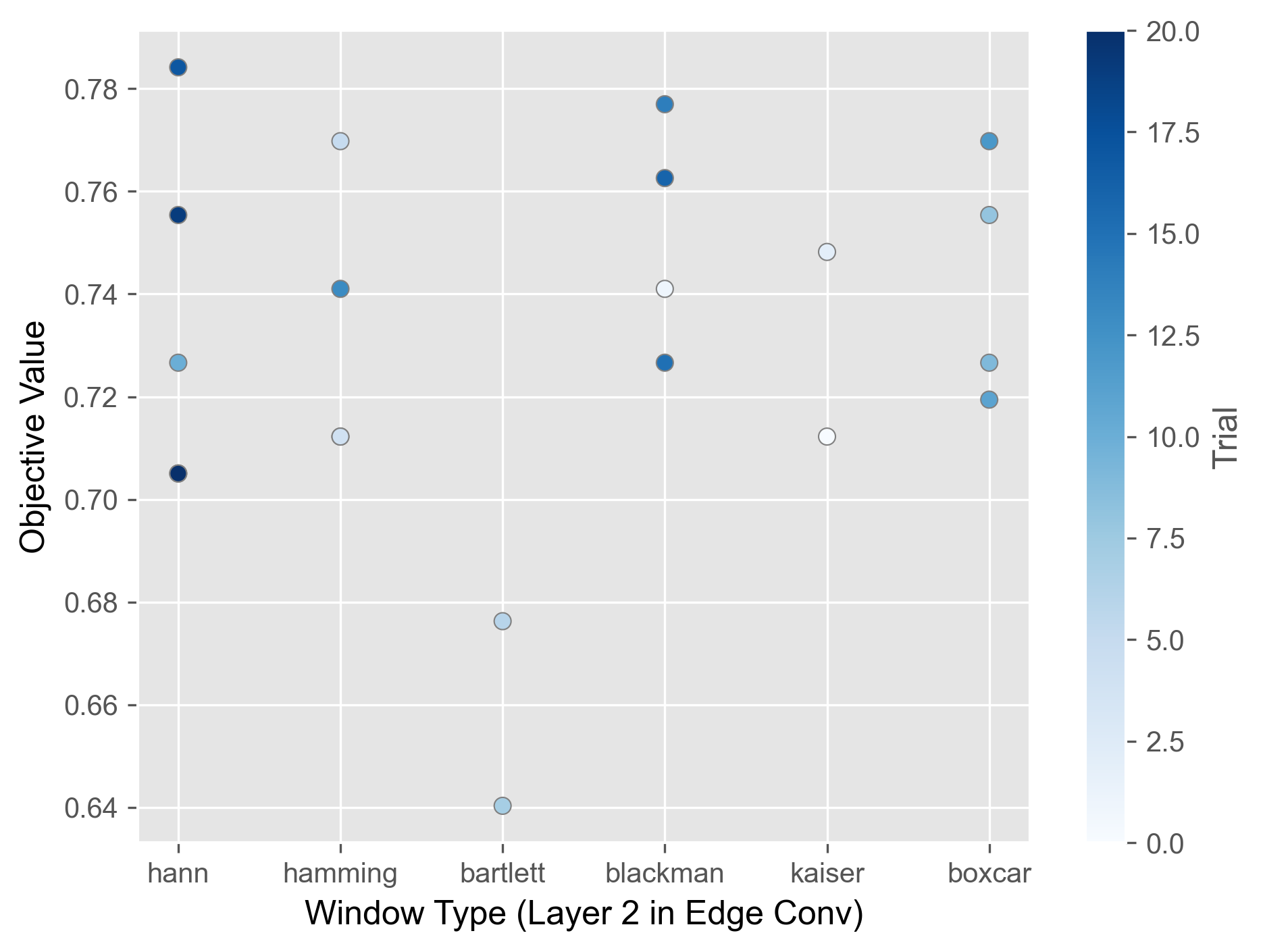}
		\caption{Effect of window types on the performance of STFT-KAN}
		\label{fig:slice_plot}
	\end{subfigure}
	\caption{Bayesian optimization history and window type impact on the STFT-KAN-based liteDGCNN}
\end{figure}

\begin{figure}[htbp]
	\centering
	\includegraphics[width=0.55\textwidth]{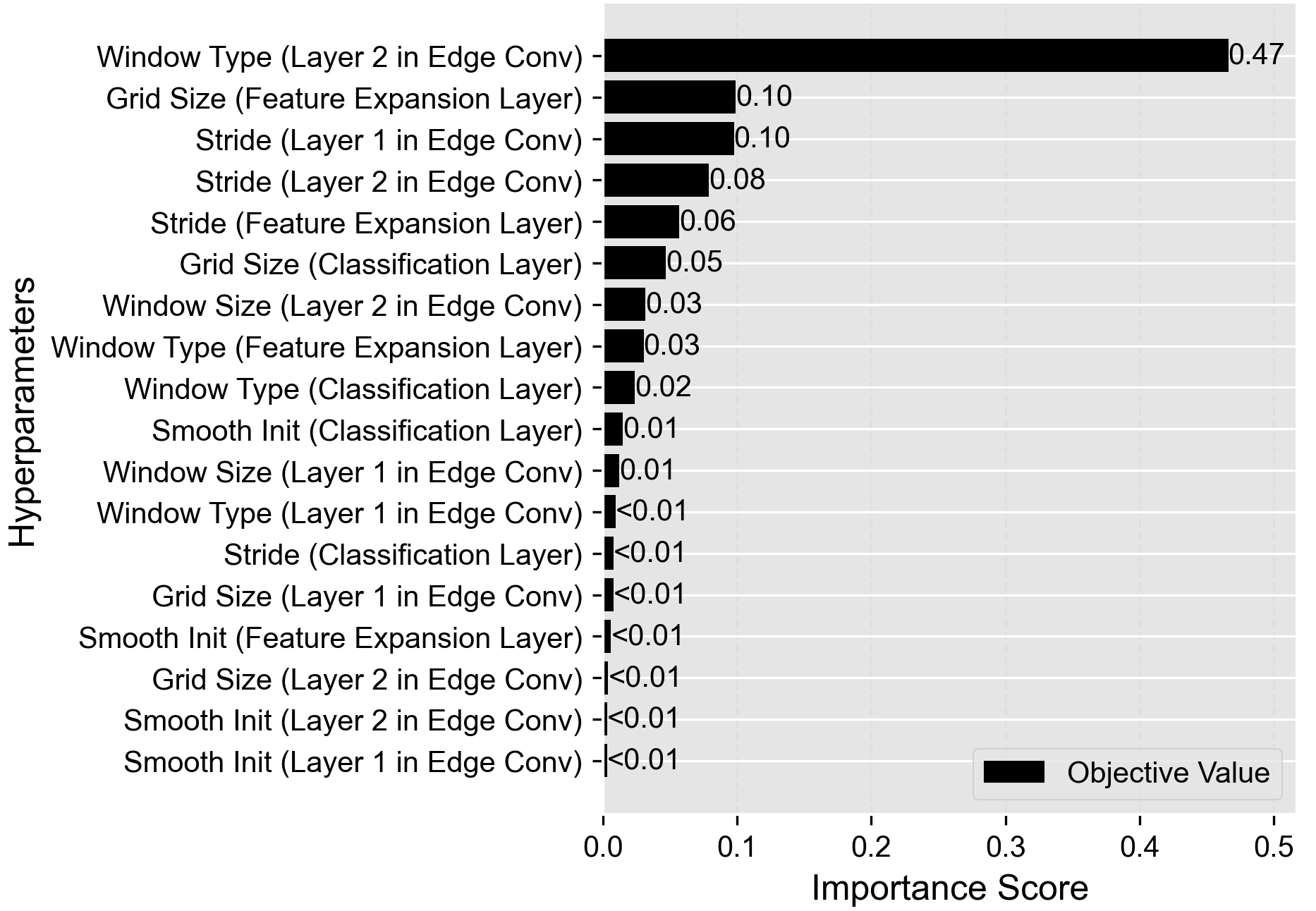}
	\caption{Parameter importances of STFT-KAN in liteDGCNN}
	\label{fig:param-importances}
\end{figure}

Bayesian optimization of the STFT-KAN-based LiteDGCNN for 3D point cloud classification of tree species resulted in the best model during trial 17, achieving an overall accuracy of 78\%. This is an improvement over previous experiments, which reached 77\%. Figure~\ref{fig:opt_history} shows the optimization history. Some trials (e.g., points 6, 7, 18, and 20) had lower objective values. Figure~\ref{fig:param-importances} highlights the importance of all parameters, while Figure~\ref{fig:slice_plot} shows that the window type in the second STFT-KAN layer used in the Edge Conv layer had the most influence on model performance, with Hann and Blackman windows providing the best results.

We focused on the four most influential parameters identified in the analysis: the window type in the second STFT-KAN layer of the Edge Conv layer, grid size of the feature expansion STFT-KAN layer, and stride values in the first and second STFT-KAN layers of the Edge Conv layer.

\begin{figure}[H]
	\centering
	\begin{subfigure}[t]{0.32\textwidth}
		\centering
		\includegraphics[width=\textwidth]{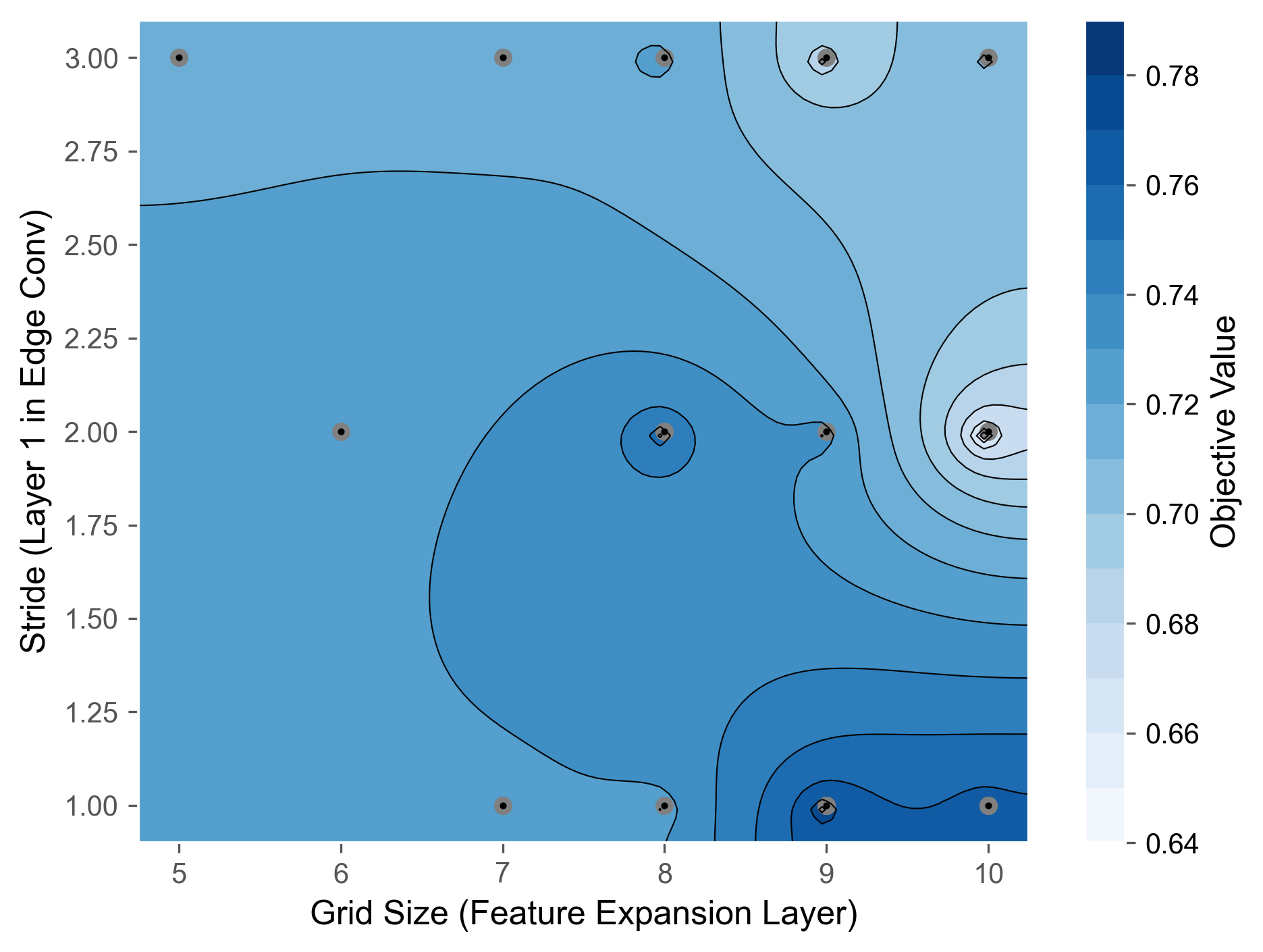}
         \caption{}
		\label{fig:contour-a}
	\end{subfigure}
	\hfill
	\begin{subfigure}[t]{0.32\textwidth}
		\centering
		\includegraphics[width=\textwidth]{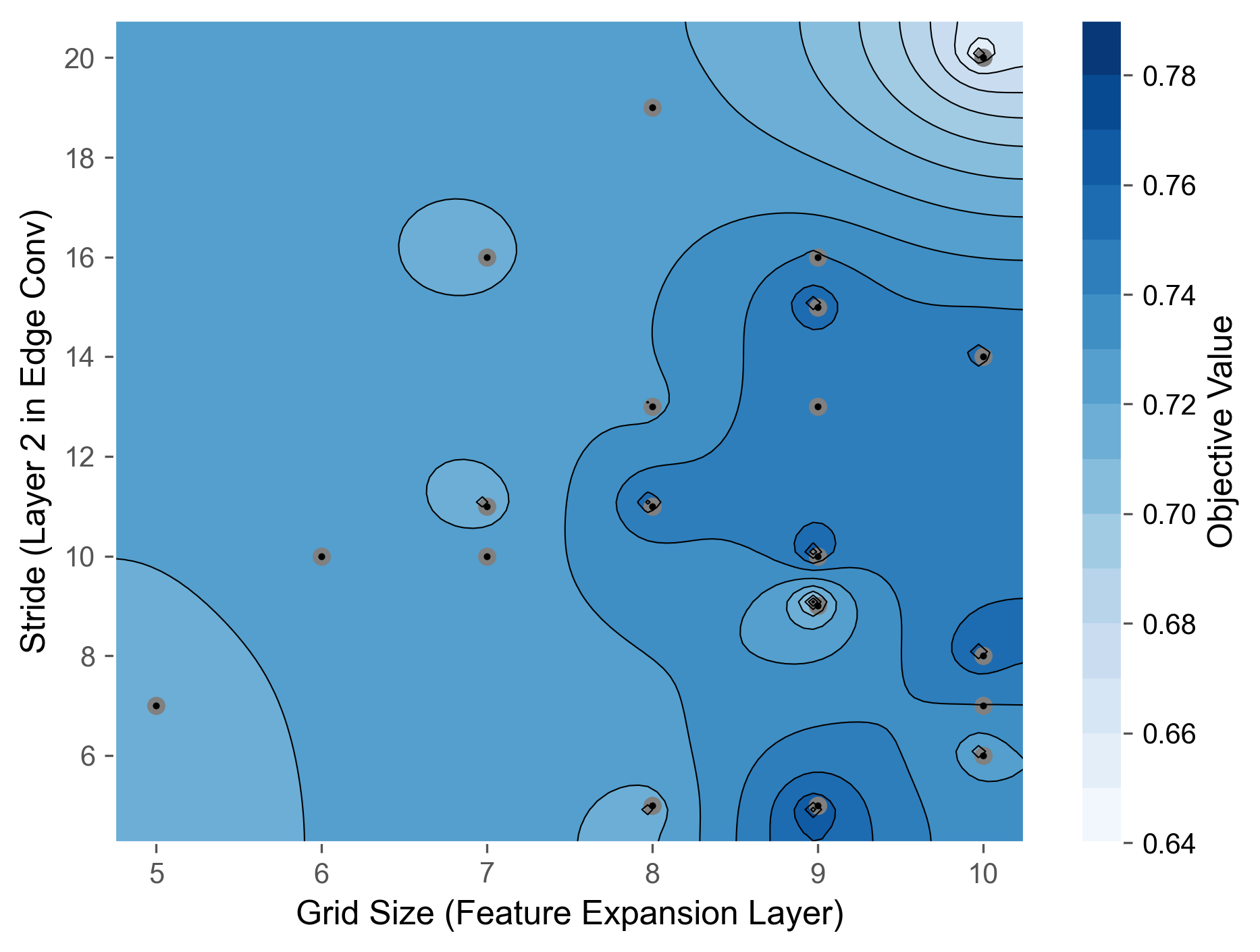}
		\caption{}
		\label{fig:contour-b}
	\end{subfigure}
	\hfill
	\begin{subfigure}[t]{0.32\textwidth}
		\centering
		\includegraphics[width=\textwidth]{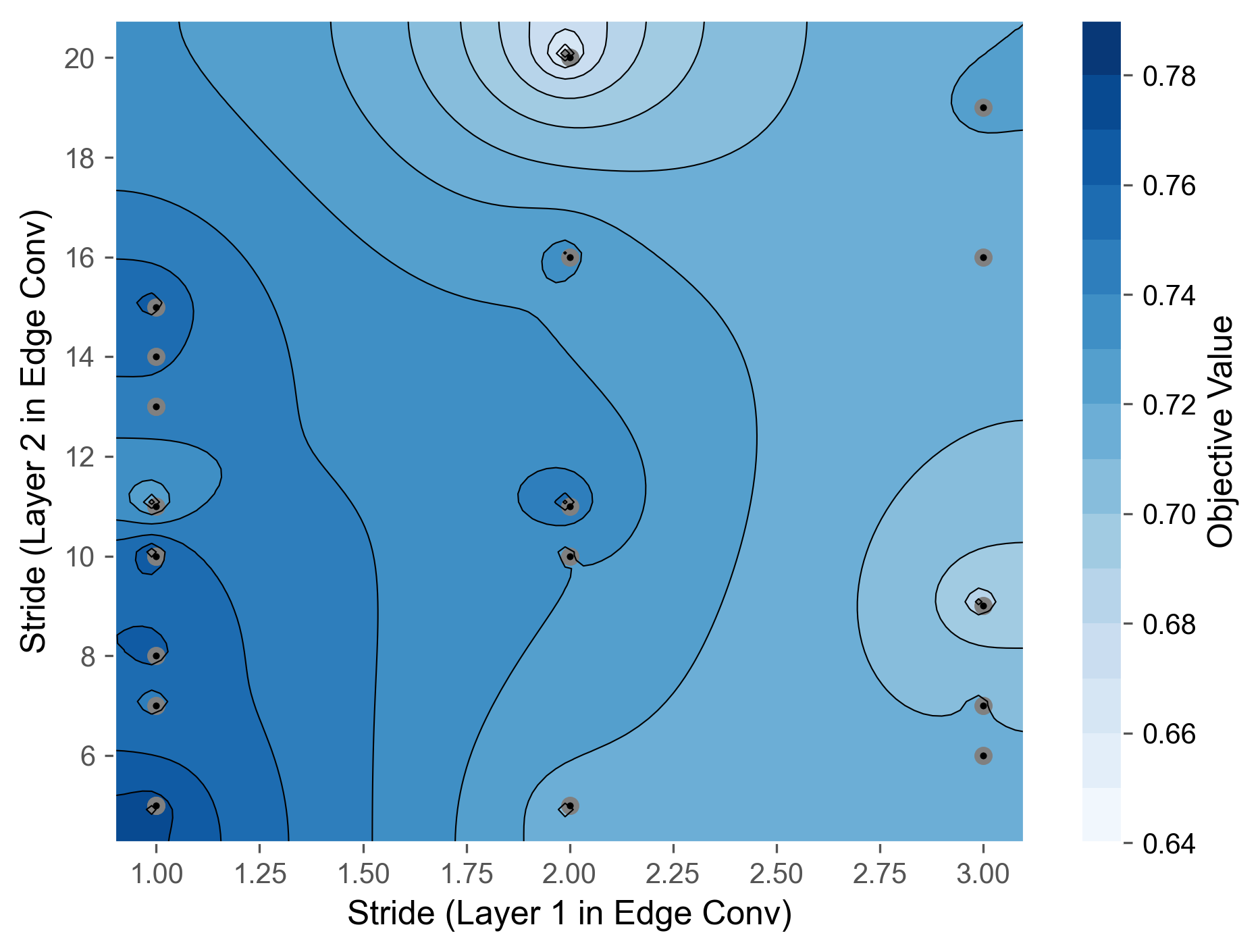}
		\caption{}
		\label{fig:contour-c}
	\end{subfigure}
	
	\vspace{1em}
	
	\begin{subfigure}[t]{0.32\textwidth}
		\centering
		\includegraphics[width=\textwidth]{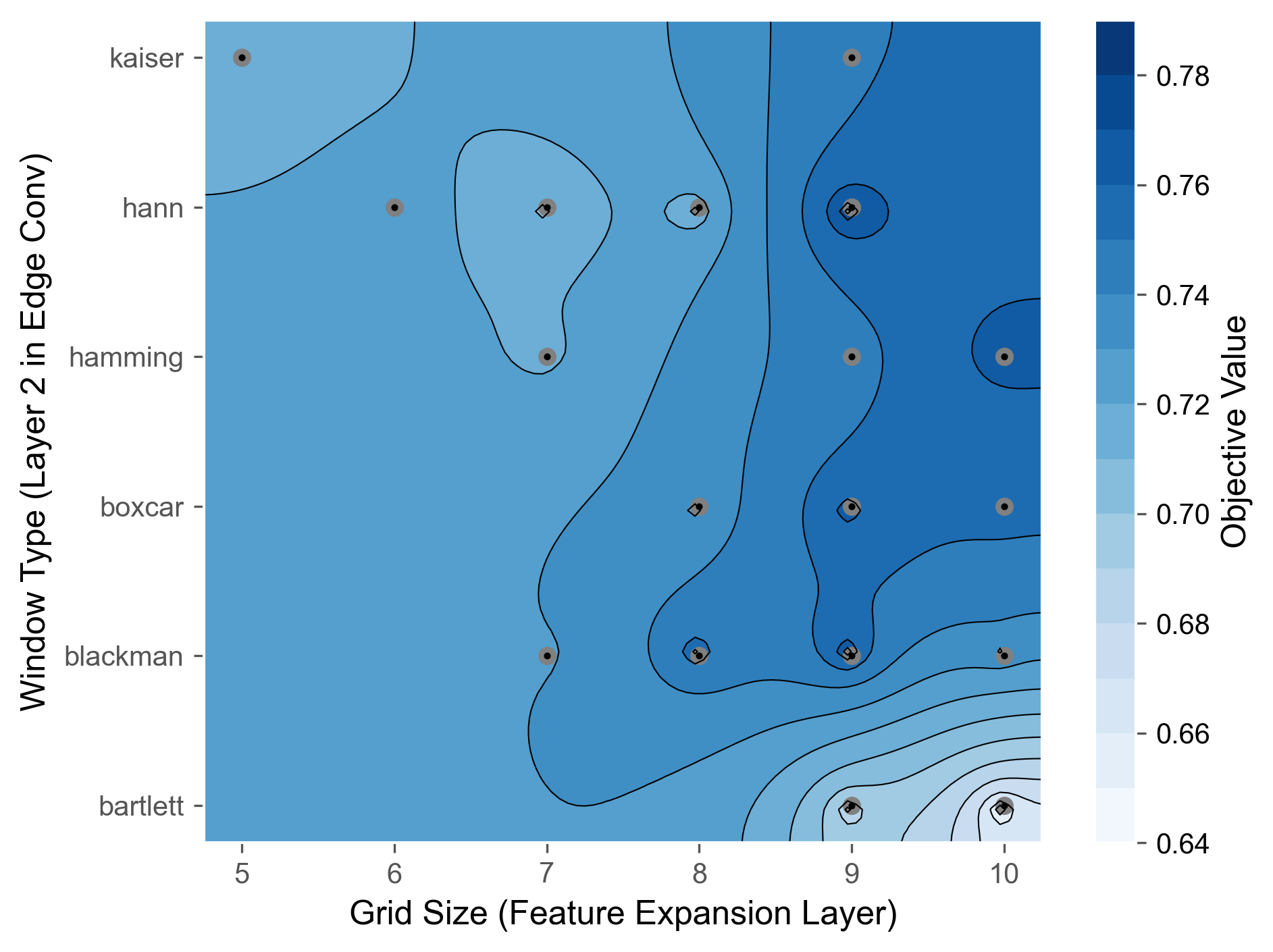}
		\caption{}
		\label{fig:contour-d}
	\end{subfigure}
	\hfill
	\begin{subfigure}[t]{0.32\textwidth}
		\centering
		\includegraphics[width=\textwidth]{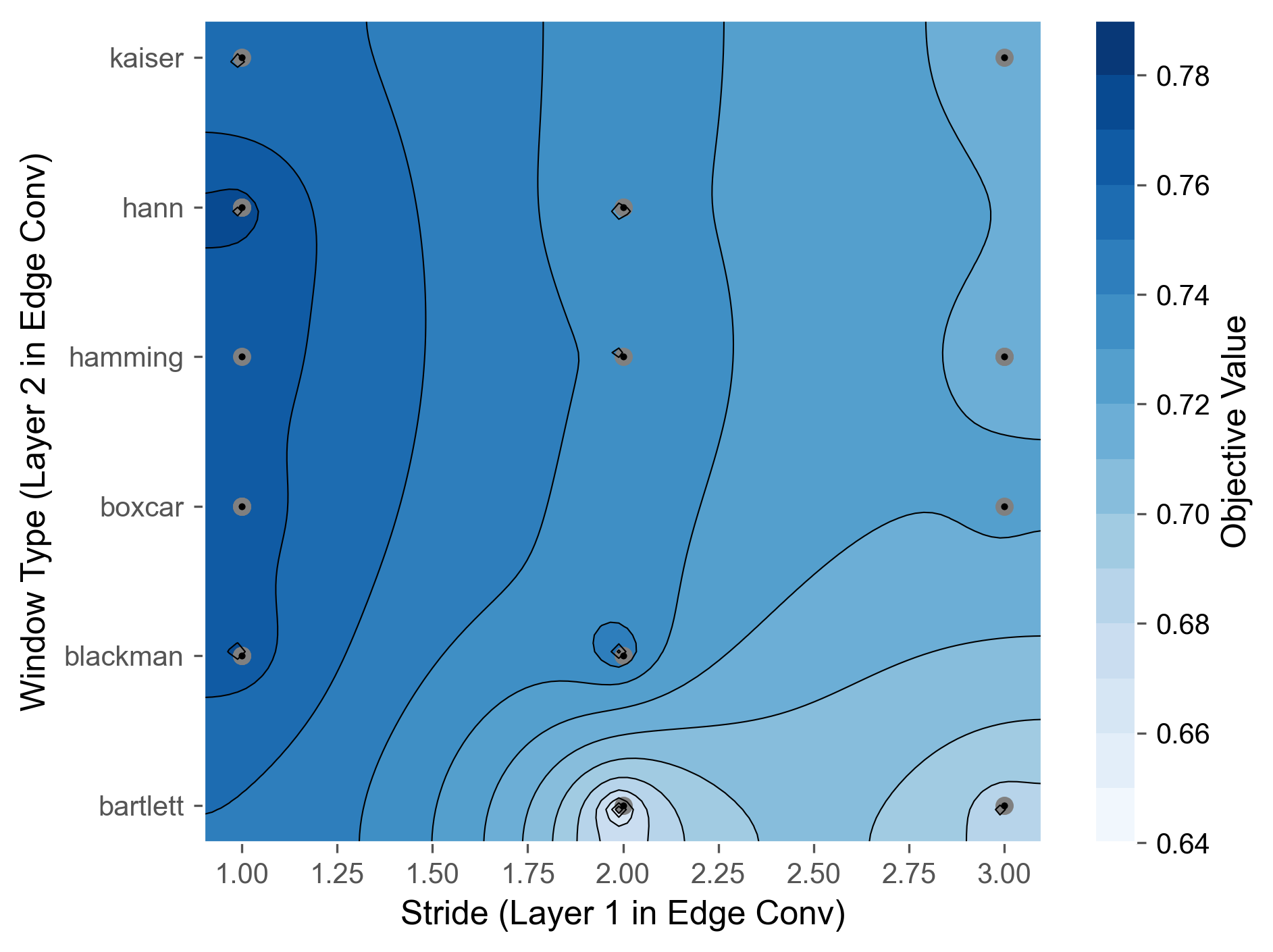}
         \caption{}
		\label{fig:contour-e}
		
	\end{subfigure}
	\hfill
	\begin{subfigure}[t]{0.32\textwidth}
		\centering
		\includegraphics[width=\textwidth]{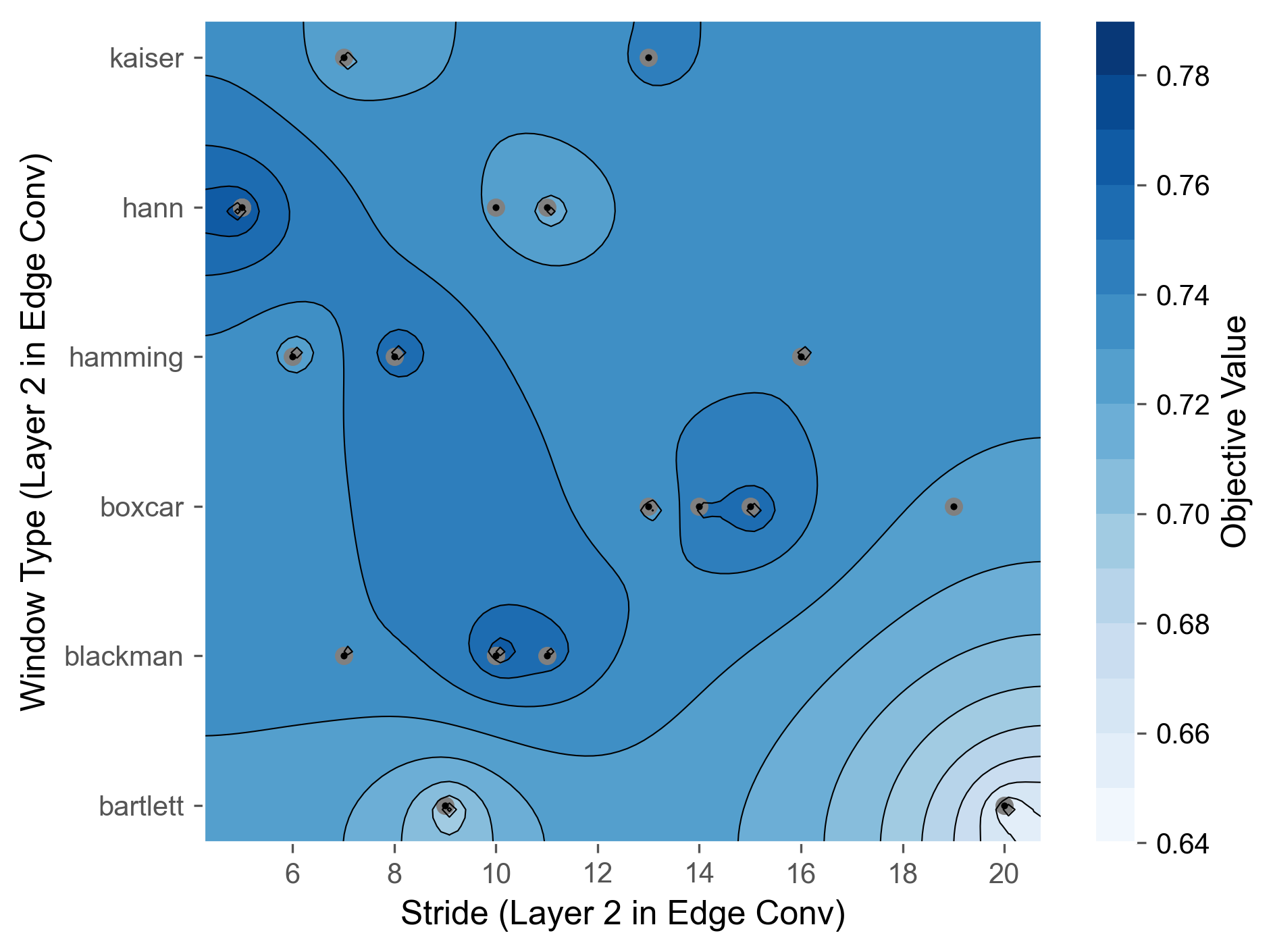}
		\caption{}
		\label{fig:contour-f}
	\end{subfigure}
	
	\caption{Contour visualization of the most important Layer Parameter combinations in STFT-KAN-based LiteDGCNN: Exploring grid size, stride, and window-type interactions.}
\end{figure}

The contour plot in Figure~\ref{fig:contour-a} shows the relationship between the grid size in the feature expansion STFT-KAN layer and stride in the first STFT-KAN layer of the Edge Conv layer. The best performance was achieved with a grid size of 9 and a stride of 1. Lower stride values generally yield better results. As the grid size increased, the performance improved up to a certain point, after which it stabilized. Figure~\ref{fig:contour-b} shows the relationship between the grid size in the feature expansion STFT-KAN layer and stride in the second STFT-KAN layer of the Edge Conv layer. We identified multiple favorable regions for the stride parameter with the best results at stride values of 5, 9, and 15. The optimization confirmed that stride five was the optimal value. The performance decreased for both larger and smaller stride values.

Figure~\ref{fig:contour-c} shows the relationship between the stride values in the two STFT-KAN layers used in the Edge Conv layer. The results showed that low stride values (1 and 2) in the first layer and moderate stride values (4, 7, 10, or 15) in the second layer provided the best results. Figure~\ref{fig:contour-d} reveals that Hann, Boxcar, and Blackman windows generally outperformed Kaiser and Bartlett windows in the second STFT-KAN layer used in the Edge Conv layer. For large grid sizes in the feature expansion layer, the Hamming window generally provided better performance, whereas for small grid sizes, the Blackman and Boxcar windows performed better.

Figure~\ref{fig:contour-e} shows the relationship between the window type used in the second STFT-KAN layer of Edge Conv and the stride of the first layer. Most window types (Kaiser, Hann, Hamming, Boxcar, Blackman) performed better than Bartlett windows, which exhibited notably lower performance, especially at medium-to-high stride values. As the stride increased, the boxcar appeared to be the best. Finally, Figure~\ref{fig:contour-f} highlights the importance of selecting an appropriate stride-window combination within the second STFT-KAN of Edge Conv. The Hann window performed best with strides 5 and 7. For strides in the range of 8-12, the Blackman window performed the best. For larger strides (14-20), the Boxcar window exhibited the best performance.

\subsection{Discussion}
The results show that the STFT-KAN and STFT-KAN-MLP models are effective for classifying 3D point clouds of tree species by adjusting the window size and stride according to the Short-Time Fourier Transform (STFT) principle. By controlling these settings, the models can capture important frequency information from the data, while reducing the number of calculations and parameters required. The STFT method allows models to analyze different frequencies in smaller time segments, making them more efficient in terms of parameters without losing essential information from the point cloud data. This helps the models to maintain good performance while being less complex.

Indeed, in STFT-KAN, identifying optimal combinations of window size, grid size, stride, and window type provides valuable insights for improving model efficiency and accuracy. In general, in architectures such as LiteDGCNN, if the input length for the STFT-KAN layer is relatively small, lower stride values for this layer tend to provide better performance. However, larger stride values are recommended for larger inputs. Increasing the grid size improves the performance, but it is important to balance the trade-off between complexity and performance to avoid using larger grid sizes that do not provide additional benefits. For window types, Blackman windows generally perform better, particularly for small grid sizes. Regarding the window size, opting for larger window sizes is advisable for larger inputs, but reductions may be necessary if the expected performance is not achieved. Bayesian optimization is a good choice for determining the best combination of parameters after defining an initial search space that suits the desired level of complexity.
  
The superiority of STFT-KAN-MLP over STFT-KAN in terms of accuracy demonstrates the strengths of MLP in feature extraction and STFT-KAN in grouping and classifying features.

STFT-KAN offers valuable options for tasks requiring efficient 3D point cloud classification, such as forest monitoring or biodiversity assessments, where real-time processing is required on mobile or embedded systems. However, the models face some limitations, particularly in terms of execution time, as the current PyTorch CUDA implementations are not fully optimized for a learnable STFT.

\section{Conclusion and perspectives}
\label{sec:conclusion}
In this paper, we introduce STFT-KAN, a new Kolmogorov-Arnold network based on the Short-Time Fourier Transform (STFT). We applied it to liteDGCNN, a lighter version of DGCNN, for tree species classification using TLS data. Our findings show that STFT-KAN offers better control over complexity, outperforming other KAN-based architectures, while delivering competitive performance to MLP. In addition, we evaluated a combined architecture that utilizes MLP for edge convolution and STFT-KAN for other layers. This architecture achieves nearly the same performance as MLP but with approximately 50\% fewer parameters. We also compared our models with state-of-the-art models used in 3D point cloud learning and found that a minimal architecture incorporating both MLP and KAN generally produces competitive results, using 87\% fewer parameters than state-of-the-art PointMLP lite, while performing better in terms of both performance and efficiency compared to other models like PointNet and PointNet++ SSG. Therefore, our model is suitable for deployment on edge devices for tree management and forest monitoring. Although STFT-KAN provides the best complexity control, experiments have shown that it has a longer execution time. In future work, optimizing the model using CUDA and parallelizing its operations (e.g., window sliding, projections, and summation) will be crucial for real-time deployment, and exploring STFT-KAN in other machine learning tasks such as time series analysis will also be essential. Additionally, we aim to demonstrate the interpretability of STFT-KAN.

\section*{Data Availability}

The dataset STPCTLS used in this study is publicly available and can be accessed via the following link: \url{https://data.goettingen-research-online.de/dataset.xhtml?persistentId=doi:10.25625/FOHUJM}. The code and preprocessed data can be found in the GitHub repository \url{https://github.com/said-ohamouddou/STFT-KAN-liteDGCNN}.

\section*{Acknowledgement}
The authors express their gratitude to the National Center for Scientific and Technical Research in Morocco for providing financial support for this research through the PhD-Associate Scholarship – PASS program.

\section*{Declaration of Generative AI and AI-Assisted Technologies in the Writing Process}

During the preparation of this work, the author(s) used ChatGPT and Claude AI to enhance readability and improve language. After using these tools/services, the author(s) reviewed and edited the content as needed and take(s) full responsibility for the content of the publication.

\bibliographystyle{elsarticle-num} 
\bibliography{references} 

\end{document}